\documentclass[letterpaper]{article}
\usepackage{iccc}

\usepackage{times}
\usepackage{helvet}
\usepackage{courier}
\usepackage{url}
\usepackage{graphicx}
\usepackage{float}
\usepackage{stfloats}
\usepackage{amsmath}
\usepackage{subcaption}

\title{Invisible Strings: \\Revealing Latent Dancer-to-Dancer Interactions with Graph Neural Networks}

\author{Luis Vitor Zerkowski\textsuperscript{1}, \hspace{.1cm} Zixuan Wang\textsuperscript{2}, \hspace{.1cm} Ilya Vidrin\textsuperscript{3}, \hspace{.1cm} Mariel Pettee\textsuperscript{4} \\
\begin{tabular}{c l}
    \textsuperscript{1} & University of Amsterdam \hspace{1.25cm}\texttt{luisvz@gmail.com} \\
    \textsuperscript{2} & Georgia Institute of Technology \hspace{.35cm}\texttt{zxwang9811@gmail.com} \\
    \textsuperscript{3} & Northeastern University \hspace{1.4cm} \texttt{i.vidrin@northeastern.edu} \\
    \textsuperscript{4} & Lawrence Berkeley National Lab \hspace{.1cm} \texttt{mpettee@lbl.gov}
\end{tabular}
}

\begin{document} 
\maketitle

\begin{abstract}
\begin{quote}

    Dancing in a duet often requires a heightened attunement to one's partner: their orientation in space, their momentum, and the forces they exert on you. Dance artists who work in partnered settings might have a strong embodied understanding in the moment of how their movements relate to their partner's, but typical documentation of dance fails to capture these varied and subtle relationships. Working closely with dance artists interested in deepening their understanding of partnering, we leverage Graph Neural Networks (GNNs) to highlight and interpret the intricate connections shared by two dancers. Using a video-to-3D-pose extraction pipeline, we extract 3D movements from curated videos of contemporary dance duets, apply a dedicated pre-processing to improve the reconstruction, and train a GNN to predict weighted connections between the dancers. By visualizing and interpreting the predicted relationships between the two movers, we demonstrate the potential for graph-based methods to construct alternate models of the collaborative dynamics of duets. Finally, we offer some example strategies for how to use these insights to inform a generative and co-creative studio practice.
    
\end{quote}
\end{abstract}

\section{Introduction}

    In many cultures, dance is deeply informed by interpersonal connection. The prevalence of social dances worldwide illustrates how dance can even, in certain contexts, be seen as an inherently interactive art form. The collaborative dimensions of dance, however, are less concrete than body positions or movements, and for this reason they can be more difficult to describe or study. Despite their invisible nature, these inter-dancer connections can provide an essential lens through which we can create and understand choreography involving more than one person.
    
    Moreover, in a culture that increasingly values digital renderings of ideas, there is significant potential for dancers to re-imagine how digital forms of dance can enrich, rather than flatten, their understanding of the medium. We present an artist-driven methodology that uses neural networks to investigate the invisible connections between pairs of dancers. 
    While machine learning techniques have been applied to solo dance performances, the intricate realm of partnering, which involves constant negotiation of weight shifts, synchronized gestures, and mutual influence, has remained relatively unexplored. We address this gap by using AI to capture subtle interactions between two dancers.\footnote{\footnotesize \url{https://github.com/humanai-foundation/ChoreoAI/tree/main/ChoreoAI_Duet_ChorAIgraphy_Luis_Zerkowski}}

    We employ a pose extraction system on video recordings from the Partnering Lab \cite{partneringlab}, translating 2D footage into 3D body poses in collaboration with the original dancers. This conversion enables us to track the trajectories of body joints and collect detailed data on each dancer’s movement, later used to understand their collective patterns. 

    We use Graph Neural Networks (GNNs) to propagate information through the dancers’ bodies, modeled as a set of nodes connected by edges. We also use Recurrent Neural Networks (RNNs) to handle the temporal nature of dance, allowing us to better capture the evolution of movements frame by frame. Because labeling how dancers influence each other is highly subjective, we adopted a self-supervised learning approach, optimizing for sequence reconstructions and focusing on the discovery of novel or unanticipated patterns rather than enforcing specific interactions.

    In this proof-of-concept work, we present results using only a subset of points from each dancer instead of the full dimensionality of the input data. We also validate our methodology using a particle dataset with a known latent interaction graph. We find that the GNN model is able to identify interesting patterns in between pairs of dancers that align with the dancers' own intuitions. Looking ahead, we suggest that this methodology has the potential to complement embodied dance research strategies by highlighting surprising and invisible connections, reflecting and refining the dancers' understandings of their own creative relationships. In doing so, we not only extend our understanding of how dancers and technologists can collaborate, but also offer new ways for dancers and choreographers to perceive and shape their own practice.

    \section{Related Work}

    \paragraph{Graph Structure Learning} Given our research interests in modeling the relationships between dancers’ movements, the field of graph structure learning is highly relevant. We can relate our model to the it by treating each body joint as a node and thinking of edges as capturing inter-joint (or inter-dancer) interactions. Graph Convolutional Networks (GCNs) \cite{kipf2017semisupervisedclassificationgraphconvolutional} extend convolutional operations to graph-structured data, allowing message passing between connected nodes. For time series data like dance, Recurrent Neural Networks (RNNs), particularly Long Short-Term Memory (LSTM) networks \cite{hochreiter1997long}, can capture temporal dependencies by maintaining hidden states over frame sequences. Variants such as Graph Recurrent Neural Networks (GRNNs) \cite{Ruiz_2020,li2017gatedgraphsequenceneural} combine these ideas, propagating information through both graph edges and time.

    \paragraph{Variational Autoencoders on Graphs} To learn the underlying structure of these dancer-to-dancer interactions, we employ methods inspired by Variational Autoencoders (VAEs) \cite{kingma2022autoencodingvariationalbayes}. VAEs have proven effective in unsupervised representation learning, enabling models to discover latent factors from complex input data. Building on this foundation, Neural Relational Inference (NRI) \cite{kipf2018neural} introduces a framework specifically designed to infer interaction graphs within systems of multiple nodes. NRI parameterizes the probability of different edge types - each one representing distinct forms of interaction - and learns these latent relationships. 
    
    \paragraph{AI and Dance} Researchers have generated 2D and 3D dance movements using a variety of machine learning techniques including RNNs \cite{graves,mccormick,chorrnn,alemi,zhou,berman,markovic}, clustering strategies like Kernel Principal Component Analysis (KPCA) \cite{james,kpca}, VAEs \cite{pettee2019imitationgenerativevariationalchoreography,papillon2022pirounetcreatingdanceartistcentric}, GNNs \cite{pettee_2020}, and Transformers conditioned on musical inputs \cite{Li2021AICM,li2020learninggeneratediversedance}.
    
    The NRI framework was previously applied to dance data by \citeauthor{pettee_2020}, but this work only considered the movements of a solo dancer. By adapting NRI to choreographic duets and enhancing its capacity of temporal understanding with GRNNs, we aim to automatically detect how one dancer’s movements influence the other, ultimately revealing what the model identifies as subtle partnering dynamics that might otherwise go unnoticed.

\section{Dataset Preparation}

    This project required transforming raw video footage of dance duets into structured pose data while addressing challenges such as multi-person tracking, occlusion, and movement complexity. We detail the dataset curation, pose extraction, and post-processing steps to ensure high-quality data for analysis and model training.

\subsection{Data Collection}

    As an artist-centric effort, our project prioritized the needs and agency of our creative collaborators. We worked closely with  Dr. Ilya Vidrin, leader of the Partnering Lab \cite{partneringlab}, to develop the scope of the project and curate the data to suit our research goals. We collected video footage of movement data from duets of dancers associated with the Partnering Lab, totalizing four videos with an aggregate duration of 41 minutes and an average duration of 10 minutes and 15 seconds. Each video was filmed with a similar single-camera setup that started with each dancer facing one another on opposite sides of the screen, each grasping the other's right hand. The dancers featured in the filmed videos opted into the project and consented to our team using the data to train models designed to analyze duets for this project.
    
    The movement styles of the dancers reflect the practices of the Partnering Lab, which ``focuses on the internal, kinesthetic experience of movement that is difficult to perceive outside of experience'' \cite{partneringlab}. Targeting micro-movements and exploring partnering as an embodiment of ethical concepts including trust and care, the dancers in these duets are highly attuned to one another, often moving slowly and deliberately, exploring mutual tensions and weight with active curiosity. Notably, the Partnering Lab team leader also regularly attended technical collaboration meetings and provided essential feedback that influenced the analysis design throughout the entire project duration.

\subsection{Pose Extraction Pipeline}

\subsubsection{2D Pose Extraction}

\begin{figure}[h]
        \centering
        \includegraphics[width=\linewidth, height=5cm]{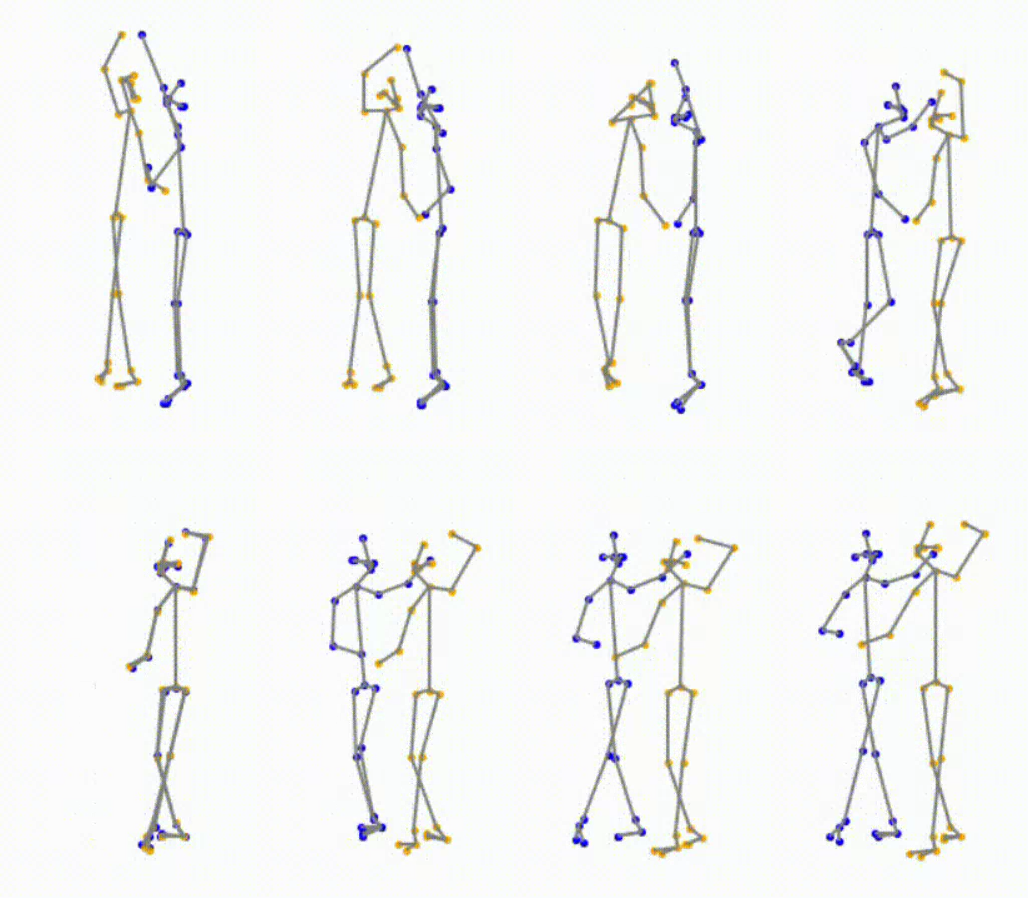}
        \caption{A raw 2D pose sequence from the Halpe pretrained model (26 keypoints) that shows vertex noise and even a missing dancer at one timestep.}
        \label{2d_halpe_grid}
    \end{figure}

    We initially experimented with 2D pose estimation models from \textit{AlphaPose} \cite{alphaposegit}, but found them insufficient for capturing spatial relationships between dancers. Also, despite the recent advances in human body pose extraction from images and multi-person tracking \cite{cao2017realtime,wei2016cpm,simon2017hand,8765346,fang2017rmpe,li2019crowdpose}, issues such as frame drops, identity swaps, and joint jitter were frequent, requiring extensive post-processing. An extraction example is shown in Figure \ref{2d_halpe_grid}.

\subsubsection{3D Pose Extraction}

    Given the limitations of 2D methods, we adopted 3D pose estimation and 3D mesh reconstruction pipelines \cite{li2021hybrik,kocabas2019vibe} to preserve depth information crucial for analyzing dancer interactions. We evaluated two models:

    \begin{itemize}
        \item \textit{VIBE} \cite{vibegit}: A prominent 3D pose and shape estimation model that outputs both joint positions and mesh reconstructions.

        \item \textit{HybrIK} (integrated with \textit{AlphaPose} \cite{alphaposegit}): A hybrid analytical-neural approach that refines pose estimation by combining classical kinematics with learning-based corrections.
    \end{itemize}

\begin{figure}[ht!]
        \centering
        \begin{subfigure}{0.45\textwidth}
            \centering
            \includegraphics[width=\linewidth, height=3.5cm]{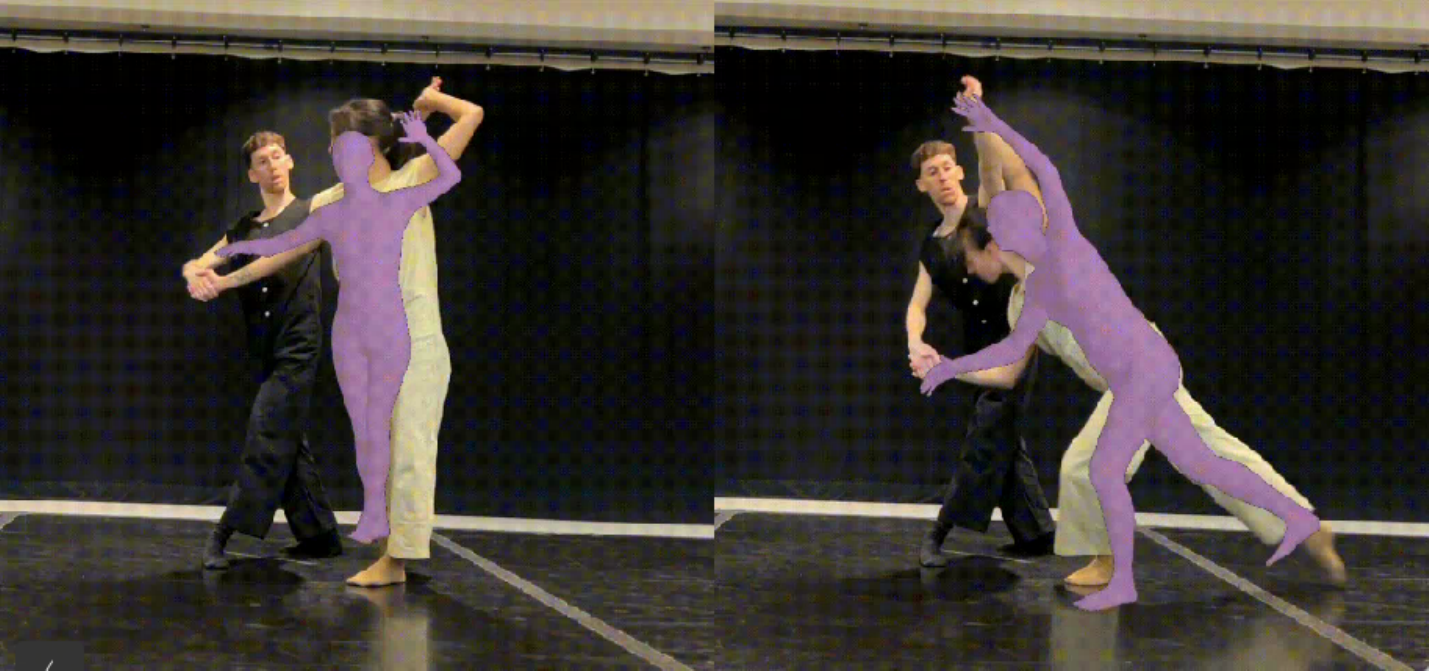}
            \caption{Mesh reconstruction from 3D pose coming from \textit{VIBE}.}
            \label{3d_vibe_mesh_grid}
        \end{subfigure}
        \vspace{0.5cm}\\
        \begin{subfigure}{0.45\textwidth}
            \centering
            \includegraphics[width=\linewidth, height=3.5cm]{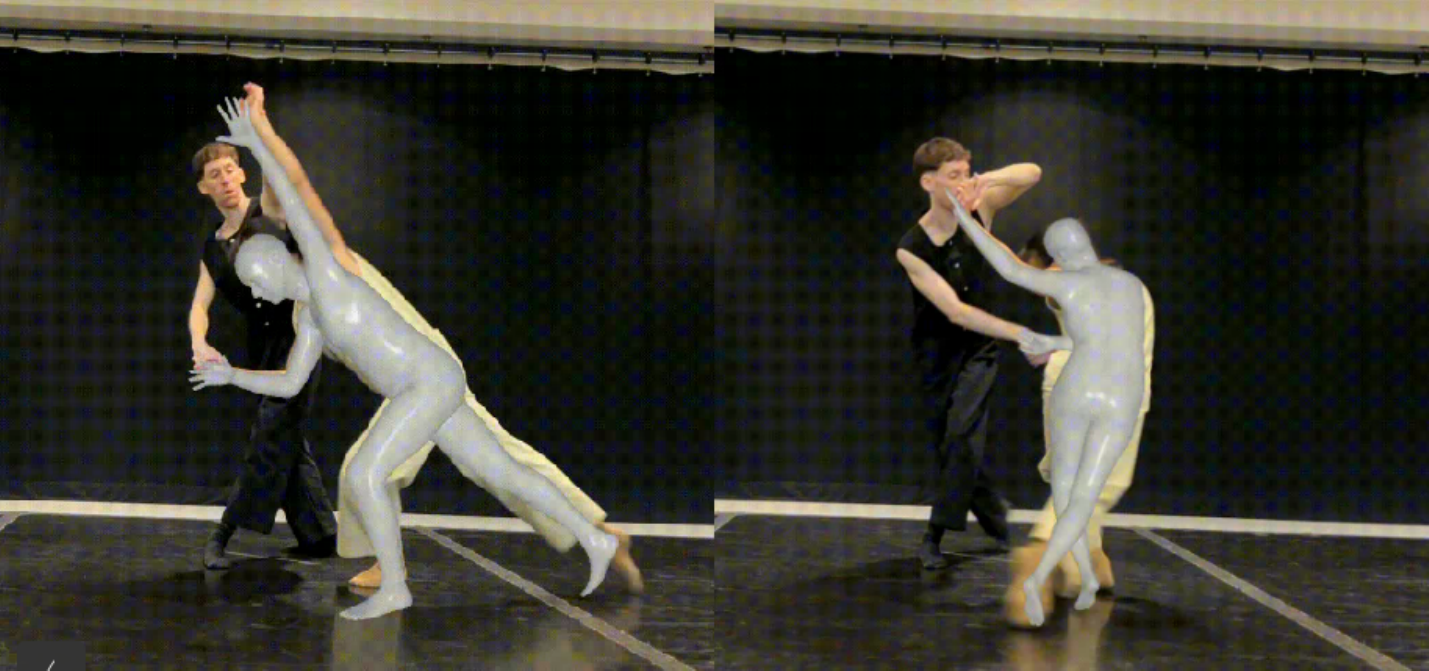}
            \caption{Mesh reconstruction from 3D pose coming from \textit{HybrIK}.}
            \label{3d_hybrik_mesh_grid}
        \end{subfigure}
        \caption{Comparison of 3D pose extractions: \textit{HybrIK} (bottom) outperforms \textit{VIBE} (top) in both simple (stationary) or complex (dynamic) movements.}
        \label{comparing_vibe_hybrik}
    \end{figure}

    In Figure \ref{comparing_vibe_hybrik}, we show an image comparison to visualize the resulting 3D joints and meshes.

    After testing, \textit{HybrIK} (via \textit{AlphaPose}) was selected for its higher pose accuracy, better multi-person tracking, and improved pose consistency. While minor noise and missing frames persisted, \textit{HybrIK}'s outputs were the most reliable for partnered dance sequences.

\subsection{Data Cleaning}

    Despite selecting the best-performing model, raw 3D pose data still contained noise and inconsistencies. To ensure stable movement trajectories, we applied several post-processing steps:

    \begin{itemize}
        
        \item \textbf{Handling Missing Frames:} If a frame had no detected poses, we duplicated the poses from the previous frame. Since our videos run at around 30 FPS, this small gap-filling strategy did not produce abrupt motion artifacts.

        \item \textbf{Frames with Single-Person Detections:} When only one dancer was detected, we compared the sum of Euclidean distances between corresponding joints for the identified person and the two people in the previous frame. We then added the person from the previous frame with the greater distance to the current frame (assuming this was the non-captured person).

        \item \textbf{Frames with More Than Two Detections:} Some frames falsely identified more than two individuals. Since our videos contained exactly two dancers, we kept only the two highest-confidence detections.

        \item \textbf{Index Consistency:} If dancers swapped IDs across frames, we scanned correct frames around the confusion frame and used the aforementioned sum of Euclidean distances between corresponding joints to correct the inversions.

        \item \textbf{Vertex Jitter:} Even successful detections contained local noise, causing “shaky” joints. We applied a 3D Discrete Cosine Transform (DCT) \cite{1672377} low-pass filter at a 25\% threshold to smooth out high-frequency jitter.
        
    \end{itemize}

\begin{figure}[ht!]
        \centering
        \begin{subfigure}{0.45\textwidth}
            \centering
            \includegraphics[width=\textwidth, height=2.5cm]{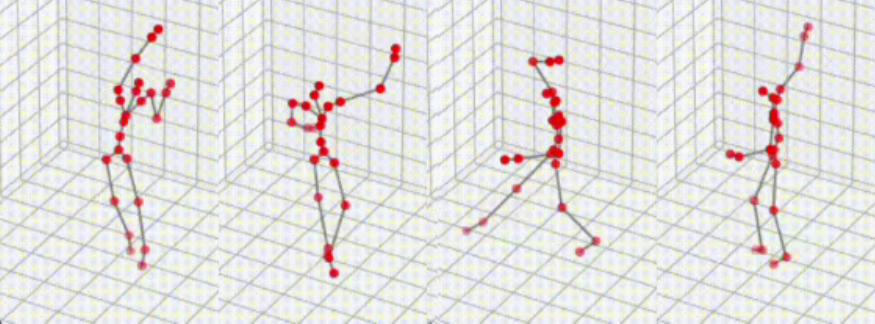}
            \caption{Pose extraction with minimal processing.}
            \label{unprocessed_pose_extraction}
        \end{subfigure}
        \vspace{0.5cm}\\
        \begin{subfigure}{0.45\textwidth}
            \centering
            \includegraphics[width=\linewidth, height=2.5cm]{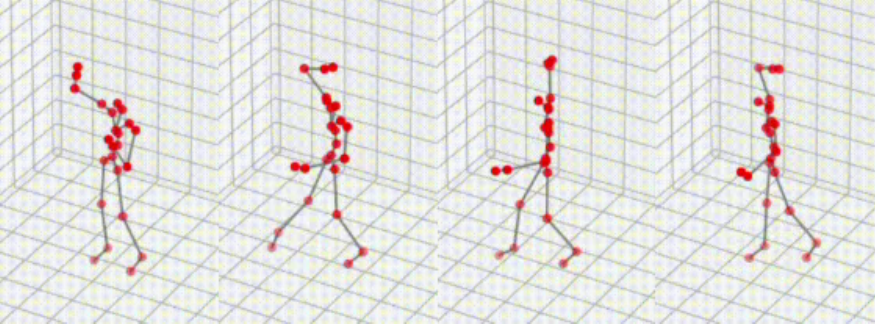}
            \caption{Pose extraction with full processing pipeline.}
            \label{processed_pose_extraction}
        \end{subfigure}
        \caption{Comparison of 3D pose extractions for a dancer: minimal processing (top) vs. full pipeline (bottom), with noticeably smoother and more realistic movements.}
        \label{comparing_unprocessed_processed}
    \end{figure}

    The impact of these corrections can be seen in Figure \ref{comparing_unprocessed_processed} by comparing the minimally-processed (i.e. handling missing frames and missing/additional people) output with our fully-processed result. The final output is cleaner, more stable, and better preserves each dancer’s bodily trajectory, which is vital for dance partnering research.

\subsection{3D Graph Construction}

    To model dancer interactions, we represented each dancer as a graph of 29 joints (nodes), a skeleton obtained from \textit{HybrIK} \cite{hybrikgit}\footnote{\footnotesize More specifically, from this issue \url{https://github.com/jeffffffli/HybrIK/issues/140} and expanded through this code \url{https://github.com/jeffffffli/HybrIK/blob/main/hybrik/datasets/h36m_smpl.py}.}, and fully connected all joints between dancers to form a dense bipartite graph. The learning objective is to classify or weight these inter-dancer edges, indicating how crucial each connection is for movement correlation.

    The data is then prepared for model training by creating batches with \textit{PyTorch} \cite{paszke2019pytorchimperativestylehighperformance} tensors. The tensors are structured with dimensions representing the total number of sequences, the sequence length, the number of joints from both dancers, and 3D coordinates + 3D velocity estimates. Finally, a 85\%-15\% training-validation split is created to allow for proper model hyperparameter tuning. To improve model generalization given our limited dataset, the training pipeline incorporates a data augmentation step that involves rotating batches of data. Each batch is rotated along the $\hat{z}$-axis by a randomly-selected angle $\theta \in [0,2\pi]$ while maintaining the original $\hat{x}$ and $\hat{y}$-axis orientations for physical consistency. This approach helps prevent the model from overfitting to the dancers' absolute positions.

    Due to the high complexity of the problem, both in the number of moving nodes and the number of edges in the graph, random joint sampling was implemented to reduce the scale of the problem. Only subsets of 6 to 10 joints (3 to 5 from each dancer) are used in each training run, and only inter-dancer edges among those sampled joints are considered. This approach helps the network converge more reliably while demonstrating proof-of-concept for larger graphs.

\section{Graph Neural Network Model}

    Understanding dancer interactions requires modeling relationships rather than individual trajectories. We extend Neural Relational Inference (NRI) \cite{kipf2018neural} to treat joints as graph nodes and infer edges representing inter-dancer connections. This section details the model’s formulation and architecture.

\subsection{Neural Relational Inference Variant}

    Our model builds on NRI \cite{kipf2018neural}, an extension of Variational Autoencoders (VAEs) \cite{kingma2022autoencodingvariationalbayes}, originally designed to infer latent interaction graphs from particle motion. By analyzing position and velocity, NRI estimates which particles influence others without predefined relationships.

    This approach is particularly relevant for duet dance analysis. Just as NRI infers particle interactions without a known ground-truth graph, we also lack a predefined interaction graph for dance partnering. The relationships between dancers shift dynamically, and there's no ground-truth connection between dancers. Instead, we employ self-supervised learning, optimizing for sequence reconstruction rather than enforcing predefined structures.

    Our model consists of an encoder and a decoder, both designed to iteratively transform node representations into edge representations and vice versa. The encoder generates edge logits, which are used to construct the latent space. Since edge indices are later sampled for the decoder, transitioning between representations is a crucial step in the process. Our model implementation includes a few important modifications from the core NRI structure \cite{kipf2018neural}:

    \begin{itemize}
        
        \item \textbf{Graph Convolutional Network (GCN):} Some linear layers are replaced with GCN layers to leverage the graph structure, improving the model's ability to capture relationships between joints. This change focuses on a subset of edges connecting both dancers rather than studying all joint relationships - as in the original implementation. Additionally, GCNs provide local feature aggregation and parameter sharing, important inductive biases for the context.

        \item \textbf{Graph Recurrent Neural Network (GRNN) Decoder:} To make better use of sequential information and potentially achieve a more suitable final embedding for predicting (or reconstructing) the next frame, one possible approach is to use a recurrent network. It is important to note that this is more of an update to a version of the NRI model rather than a completely new idea. The authors had already explored a recurrent decoder, with the main differences being the previously mentioned modifications in layers and the use of GRU nodes instead of LSTM nodes (introduced in the next item).

        \item \textbf{Custom GCN-LSTM Cells:} To utilize the recurrent structure crucial for sequence processing while maintaining graph information and GNN architecture, the classic LSTM cell has been reimplemented with GCN nodes. In the final version of our architecture, only the decoder incorporates the recurrent component, which generates a final sequence embedding that the model uses to reconstruct the next frame.
        
    \end{itemize}

    By incorporating these modifications, the model maintains the core principles of the original NRI while theoretically enhancing its ability to generalize and adapt to the dynamics of ``connected" particles moving. We now describe the architecture (shown in figure \ref{final_arch}) more precisely:

    \begin{figure*}[h]
        \centering
        \includegraphics[width=\textwidth]{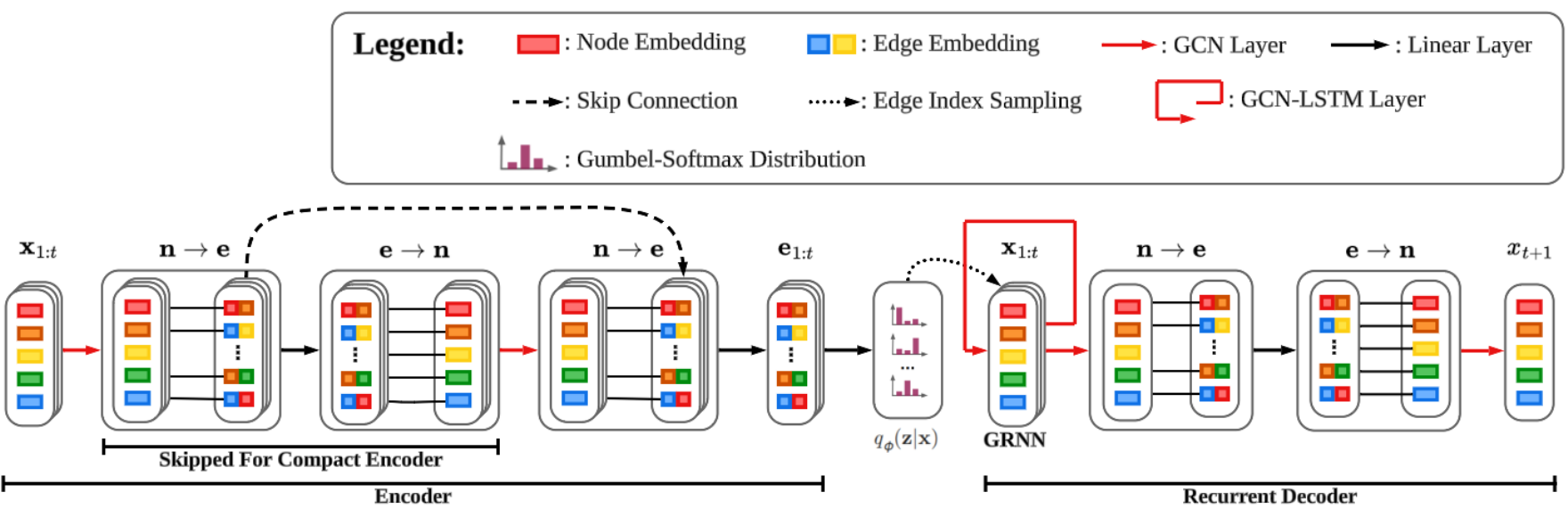}
        \caption{Schematic of the final model architecture, including the GCN nodes and the GRNN adapatation, inspired by the one found in the original NRI paper \cite{kipf2018neural} (Figure 3, page 3).}
        \label{final_arch}
    \end{figure*}

\subsubsection{Encoder}

    \begin{itemize}
        
        \item \textbf{Node-to-Edge Transformation:} Dancer joints are passed through a GCN layer with 64 latent dimensions and then converted to edge representations, which doubles this representation.

        \item  \textbf{Linear Layer + Batch Normalization + Dropout:} A linear layer refines edge embeddings, receiving input of dimension 128 dimensions and generating an output of 64 dimensions. Optionally we use batch normalization and dropout with 10\% probability  for regularization.

        \item \textbf{Edge-to-Node Transformation:} The edge representations are converted back into nodes, and another GCN layer with same dimensionality as before is applied.

        \item \textbf{Second Node-to-Edge Transformation:} The nodes are then transformed back into edges, followed by another Linear layer from 192 dimensions (this time we have 3$\times$64=192 because we also use a skip connection coming from the first Node-to-Edge transformation) to 64 dimensions.

        \item \textbf{Logits Computation:} A final linear layer outputs logits for each possible edge type. For this layer, we mainly tested with binary edges ("existing", "non-existing") or 3 edge types (“no connection,” “weak” and “strong”), but we also added the possibility of 4 or even 5 edge types. Each one of these options comes with a prior probability distribution that guides the latent distribution, making the later sampling process gravitate towards a certain allocation of each edge type. 
        
\end{itemize}

\subsubsection{Decoder}

    \begin{itemize}
        
        \item \textbf{Edge Index Sampling:} The decoder begins by hard sampling edge indices using a Gumbel-Softmax distribution \cite{maddison2017concretedistributioncontinuousrelaxation,jang2017categoricalreparameterizationgumbelsoftmax}, based on the logits generated by the encoder and the prior probabilities of each edge type. While this process results in discrete edges (i.e., edges are either present or absent, without intermediate probabilities), it effectively approximates sampling from a continuous distribution and uses Softmax to enable the reparameterization trick, keeping the entire pipeline fully differentiable.

        \item \textbf{GRNN + Node-to-Edge Transformation:} Once edges have been sampled, the decoder processes the data through a GRNN composed of modified LSTM nodes with GCN layers, followed by a transformation of the final sequence embedding into edge representations with 64 dimensions again.

        \item \textbf{Linear Layer + Batch Normalization + Dropout:} A linear layer refines edge embeddings, receiving input of dimension 128 dimensions and generating an output of 64 dimensions, and optionally using batch normalization and dropout with 10\% probability for regularization.

        \item \textbf{Edge-to-Node Transformation:} The edge representations are converted back into nodes, and a GCN layer is applied to predict (or reconstruct) the next frame of the input sequence.
        
    \end{itemize}

\section{Experiments and Results}

    We evaluated our model on two datasets: a charged-particle dataset from the original NRI paper, serving as a simpler dataset with a known interaction graph to help benchmark our model performance, and the duet dance dataset, featuring real-world 3D pose sequences with two dancers at a time.

\subsection{Charged Particle $n$-Body Simulations}

    To validate the architecture’s viability under more controlled conditions, we used the original charged-particle dataset used in \citeauthor{kipf2018neural}. We generated 50,000 simulated trajectories of five particles each, with 2D positions and velocities over 49 frames. A random interaction graph dictates repulsive or attractive forces between particles.

\begin{figure}[h]
        \centering
        \includegraphics[width=0.45\textwidth, height=4cm]{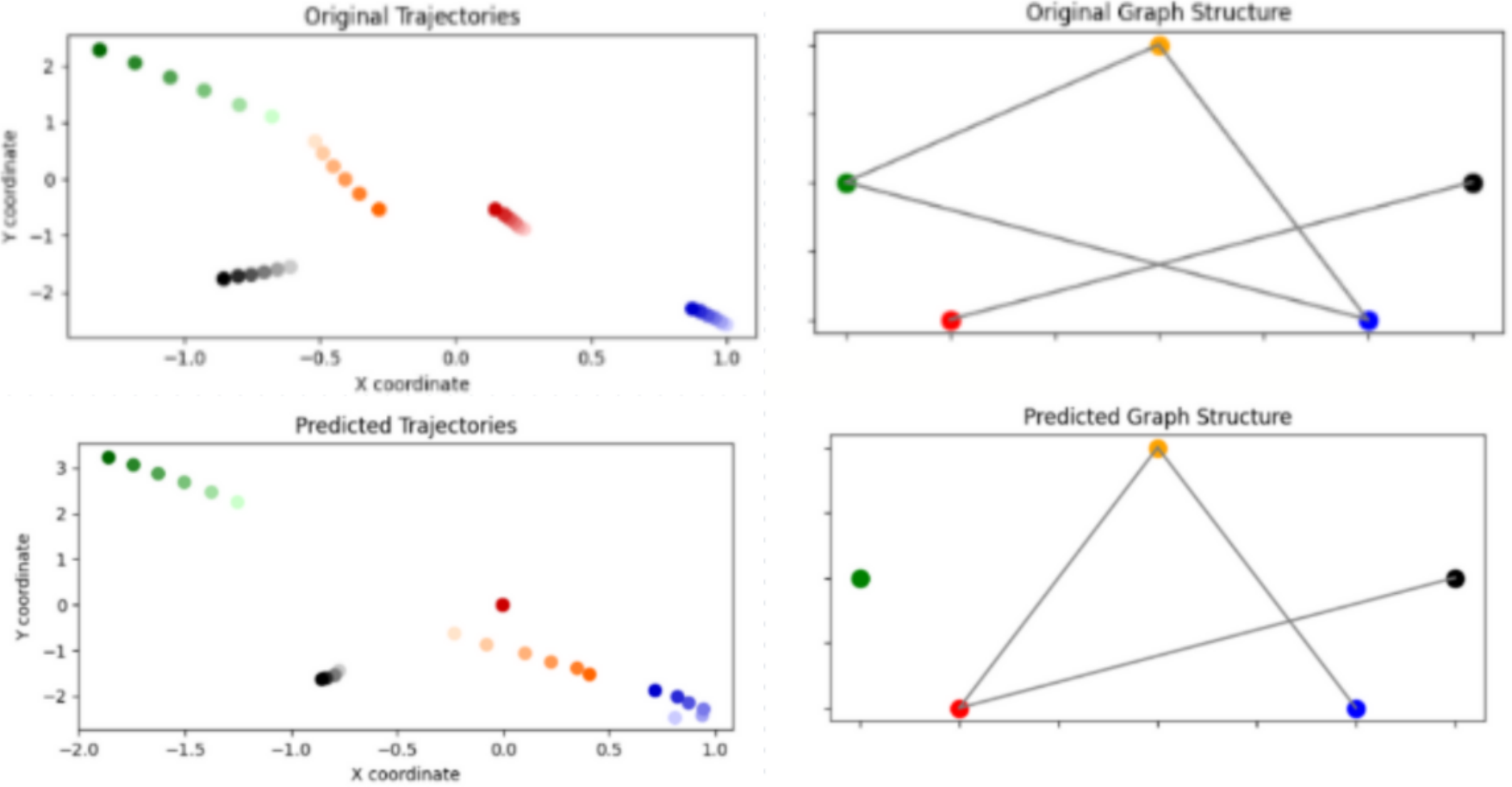}
        \caption{On top, original simulated trajectories and original sampled edges. On bottom, reconstruction and edge prediction results. The model accurately captured the movement and location of three particles (green, black, blue), approximated movement shape for one (orange) despite location inaccuracies, and positioned the last (red) reasonably well but without movement.}
        \label{trajectories_and_graph}
    \end{figure}

    We tested a variety of architectures, and the results were promising. Figure \ref{trajectories_and_graph} presents an example from one of our models, showing that the predicted graph structures and particle trajectories generally aligned with the ground truth. Additional validation results can be found in our open-source codebase.

\subsection{3D Dance Duets}

    To assess model performance on the 3D dance data, we use predicted movement fidelity as a proxy for evaluating sampled edges, given that the subjectivity of defining correct inter-dancer connections makes defining true edge labels difficult. Models were trained until convergence, typically for 15 to 20 epochs, and resulted in stable reconstruction loss curves with no signs of overfitting on validation data.

    \begin{table}[ht]
        \centering
        \caption{\textbf{Reconstruction Mean Squared Error} for multiple tasks, input sequence lengths ($\ell$), number of edge types ($n$), and model architecture configurations (compact encoder vs. full architecture as introduced in Figure \ref{final_arch}). Bold highlights the best (smallest) reconstruction error.}
        \label{tab:mse_values}
        \begin{tabular}{ccccc}
        \hline
        \textbf{Task} & \textbf{$\ell$} & \textbf{$n$} & \textbf{Model} & \textbf{MSE} \\
        \hline
        5-Body Sim. & 6 & 3 & Compact Enc. & 0.70 \\
        5-Body Sim. & 6 & 3 & Full Arch. & 0.69 \\
        5-Body Sim. & 6 & 4 & Compact Enc. & 0.82 \\
        5-Body Sim. & 12 & 3 & Compact Enc. & 0.33 \\
        5-Body Sim. & 12 & 3 & Full Arch. & \textbf{0.32} \\
        6-Joint Dance & 8 & 4 & Full Arch. & 0.58 \\
        \hline
        \end{tabular}
    \end{table}

    KL-divergence loss, however, plateaued early, limiting latent space exploration. While beta coefficient scaling mitigated this effect, short training runs reduced its effectiveness. Longer training stabilized KL behavior, but these experiments were conducted on earlier, smaller models without a Graph Recurrent Neural Network (GRNN), which made extended training less feasible in the final architecture.

    For one of our best-performing models, we trained with 6 sampled joints (3 per dancer), with dancer rotation augmenting the dataset 10 times. The full encoder was used with 8-frame input sequences, 64 hidden dimensions, and 4 edge types. It was trained for 20 epochs (18 hours) using mean squared error. Table \ref{tab:mse_values} shows some MSE values for multiple trained models.

\subsubsection{Movement Predictions}

\begin{figure}[h]
        \centering
        \includegraphics[width=0.45\textwidth]{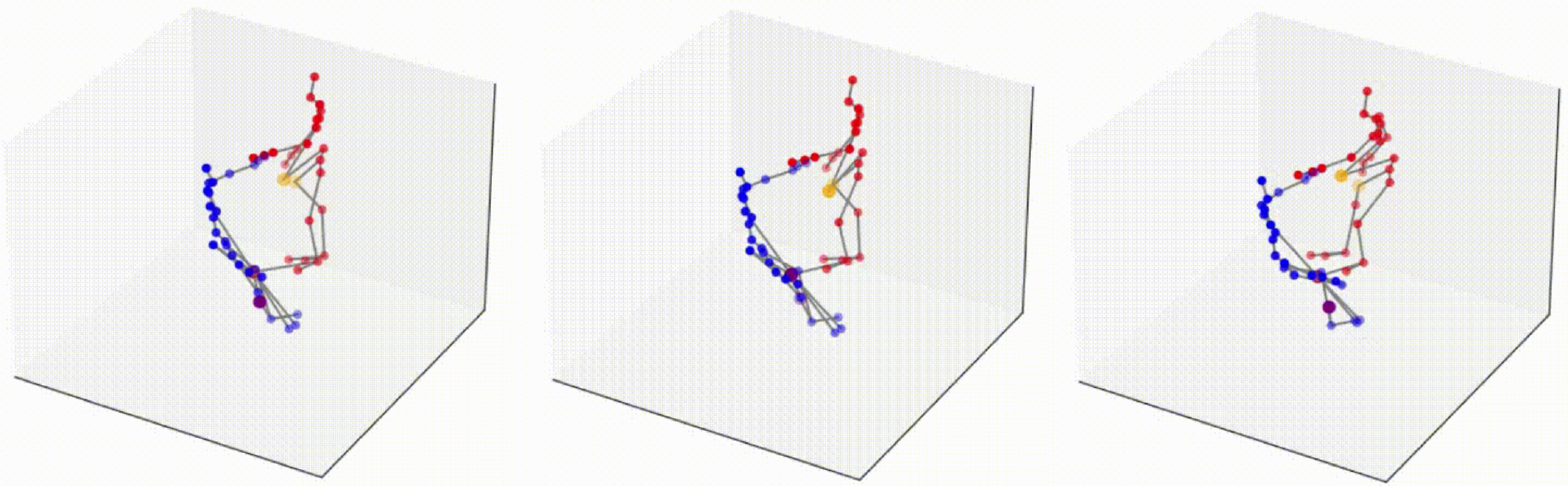}
        \caption{Example of a good reconstruction. Reconstructed sampled joints are color-coded for clarity: purple for the blue dancer and orange for the red dancer.}
        \label{good_reconstruction}
    \end{figure}

    We consider a random subset of 3-5 joints for each of the two dancers for both model training and predicted movement evaluations. The predicted movements are highly sensitive to sampled edges, as information in a graph network spreads through connected nodes. The best-performing models tend to assign multiple connections to a single joint, enabling smoother information flow between dancers. As seen in Figure \ref{good_reconstruction}, sampled joints maintain accurate spatial positioning and move coherently with the body. Given that the sampled connections only consist of joints between the dancers, it's notable that the captured interaction can lead to accurate movement predictions for each of the individual dancers.

    The reconstructed frames provide valuable insight into how the model interprets connections between particles. By analyzing the learned edge distribution and sampled edges across different examples, it becomes possible to better understand the network’s perception of dancer interactions and movement patterns.

\begin{figure}[ht!]
        \centering
        \begin{subfigure}{0.45\textwidth}
            \centering
            \includegraphics[width=0.95\textwidth, height=2.25cm]{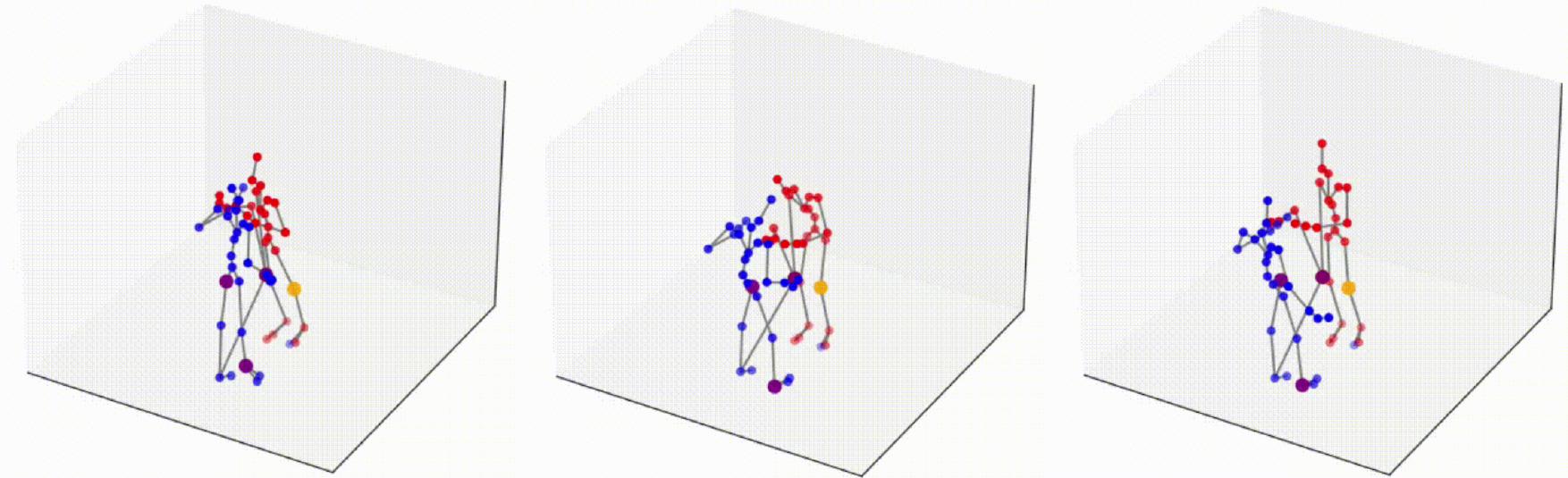}
            \caption{Example of edge sampling limiting reconstruction to 3 joints, while the other 3 remain static at the center of the coordinate frame despite being sampled.}
            \label{bad_reconstruction_1}
        \end{subfigure}
        \vspace{0.5cm}\\
        \begin{subfigure}{0.45\textwidth}
            \centering
            \includegraphics[width=0.95\linewidth, height=2.25cm]{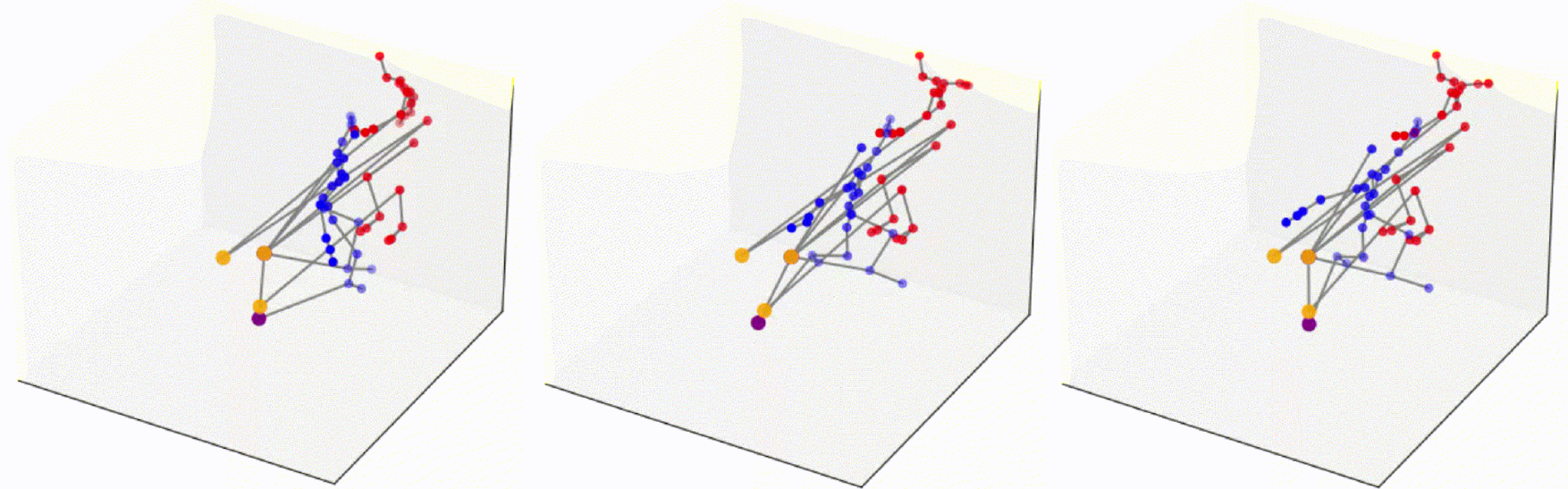}
            \caption{Example illustrating the model's reconstruction challenges when dancers are farther from the center, resulting in joints that are either stationary or inaccurately positioned.}
            \label{bad_reconstruction_2}
        \end{subfigure}
        \caption{Examples of poor reconstructions.}
        \label{bad_reconstructions}
    \end{figure}

    Certain limitations arise due to the nature of the edge sampling and model assumptions. Common sources of reconstruction errors include:

    \begin{itemize}
        
        \item \textbf{Shaking and jitter:} Some particles exhibit noticeable instability, partly due to inherent noise in the original sequence and also because each reconstructed frame is generated independently, without leveraging temporal smoothing since the model predicts only one frame at a time.

        \item \textbf{Limited movement in sampled joints:} In some cases, the sampled edges do not adequately capture movement, resulting in reconstructions in which the joints remain nearly stationary (see e.g. Figure \ref{bad_reconstructions}, top). This effect is particularly pronounced when the sampled connections fail to establish a strong information pathway between dancers.

        \item \textbf{Challenges with dancer switching:} When dancers cross paths or switch sides, the reconstructed joints sometimes remain near their initial positions rather than following the dancers' actual motion. This suggests that the model has difficulty generalizing to sequences where relative spatial relationships shift significantly.

        \item \textbf{Drift towards the center:} If the dancers move away from the origin, reconstructions tend to stay near the center (Figure \ref{bad_reconstructions}, bottom), which could be improved using a different form of pre-processing.
        
    \end{itemize}

    Regardless of these challenges, the reconstructions still generally capture meaningful aspects of the dancers’ interactions. The observed limitations highlight areas for future refinement, such as improving edge sampling strategies and incorporating temporal consistency in predictions.

\subsubsection{Predicted Edges}

    During inference for edge prediction, edges are sampled differently from the training phase. Instead of hard sampling, which enforces a strict edge selection, we use soft sampling (i.e. sampling with associated probability) to retain only high-confidence connections. This ensures that only the most structurally relevant edges remain in the final reconstruction, reducing noise and improving interpretability.

    A key observation is that the number of sampled edges with a confidence above 80\% remains consistently low and aligns closely with the prior distribution of edge types (whether two, three, or four types are used). This suggests that the learned latent space effectively captures meaningful movement relationships rather than introducing arbitrary edges. Since the edge priors were designed to reflect a sparse but essential connectivity between dancers, this alignment reinforces the idea that the model is not simply memorizing movement sequences but actively identifying the most influential joints for movement propagation.

    The predicted edges exceeding the 80\% confidence threshold have interesting several features: 
    
    \begin{itemize}
    
    \item First, most of these connections tend to have similar confidence levels. This indicates a \textbf{low hierarchy among selected edges} (Figure \ref{low_hierarchy_edges}). In other words, once an edge is classified as important, it tends to contribute equally to the reconstruction, rather than one edge being clearly dominant. This is particularly interesting from a movement analysis perspective because it suggests that information is distributed across multiple interaction points rather than concentrated in a single connection.

    \begin{figure}[h]
        \centering
        \includegraphics[width=0.45\textwidth]{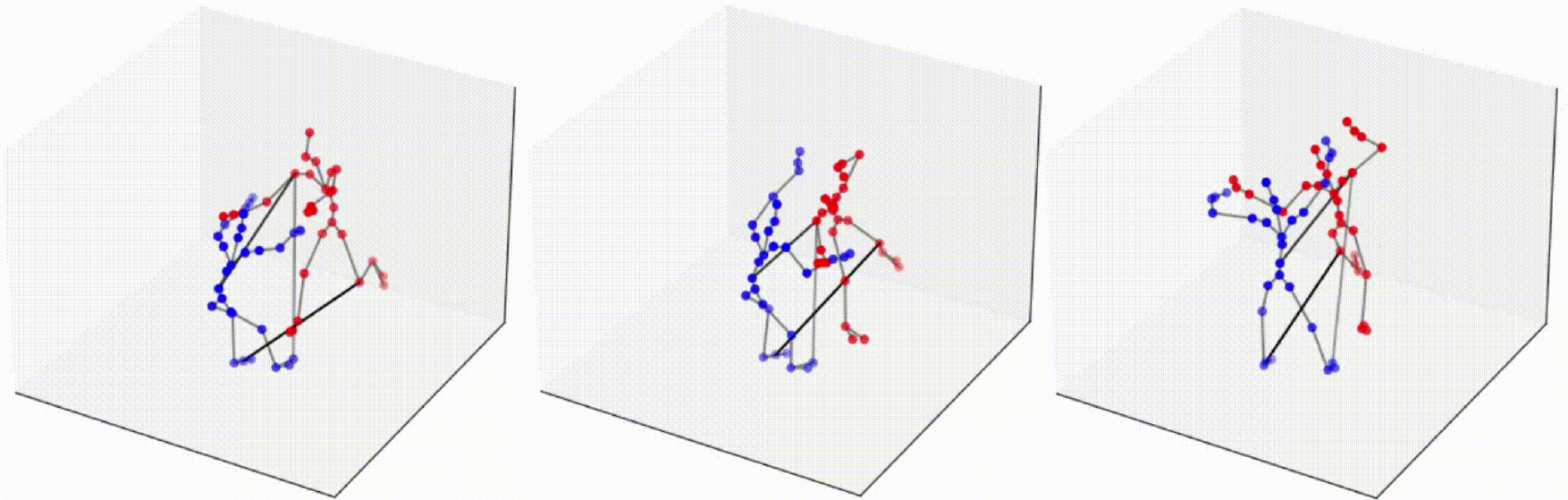}
        \caption{Example of the sampled edge distribution. The black edges represent connections between the dancers, with darker edges indicating higher confidence in their importance for reconstruction. In this typical case, 3 edges were selected for 6 sampled joints, 2 with slightly higher importance, though all exceed 80\% confidence.}
        \label{low_hierarchy_edges}
    \end{figure}

    \item Additionally, it is common for a single joint to be connected by multiple edges, meaning that \textbf{certain key joints act as hubs} in the reconstructed interaction graph (Figure \ref{multi_edges}). This aligns with prior observations that information propagates more effectively when there are multiple connected paths leading to and from the same dancer (through one joint or within two hops). For instance, a dancer’s hand might influence their partner’s torso, while simultaneously being connected to their partner’s own hand or shoulder, helping to propagate multiple layers of movement dependency.

    \begin{figure}[ht!]
        \centering
        \begin{subfigure}{0.45\textwidth}
            \centering
            \includegraphics[width=0.95\textwidth, height=2.25cm]{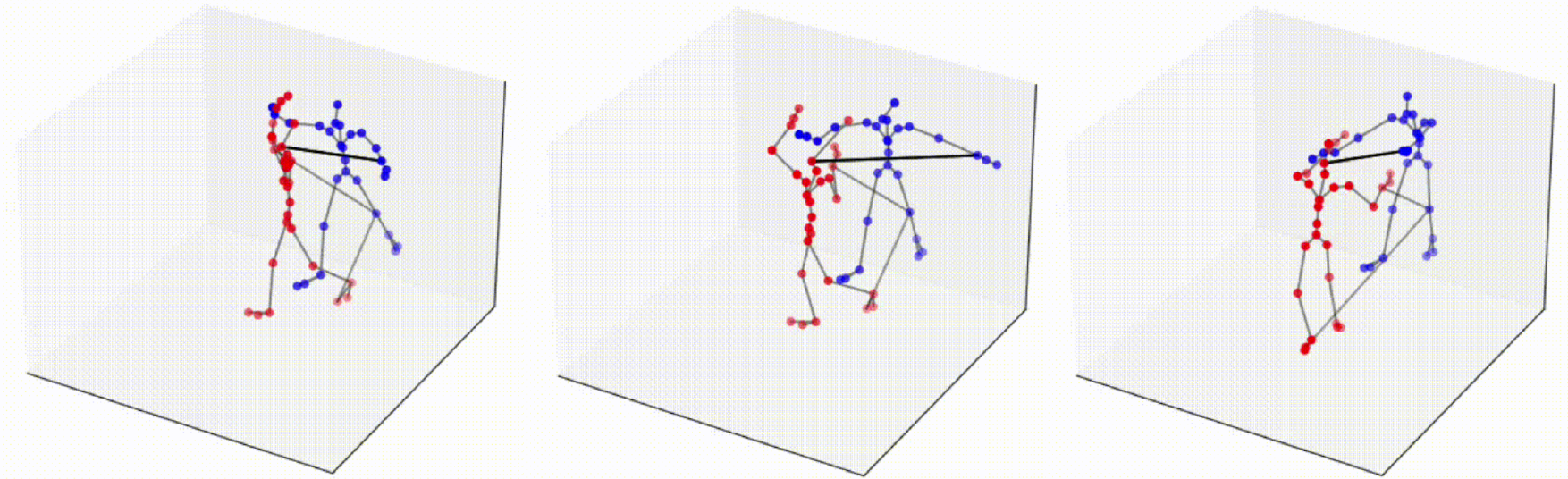}
            \caption{An undirected example illustrating simple movement for a set of 6 sampled joints.}
            \label{multi_edge_1}
        \end{subfigure}
        \vspace{0.5cm}\\
        \begin{subfigure}{0.45\textwidth}
            \centering
            \includegraphics[width=0.95\linewidth, height=2.25cm]{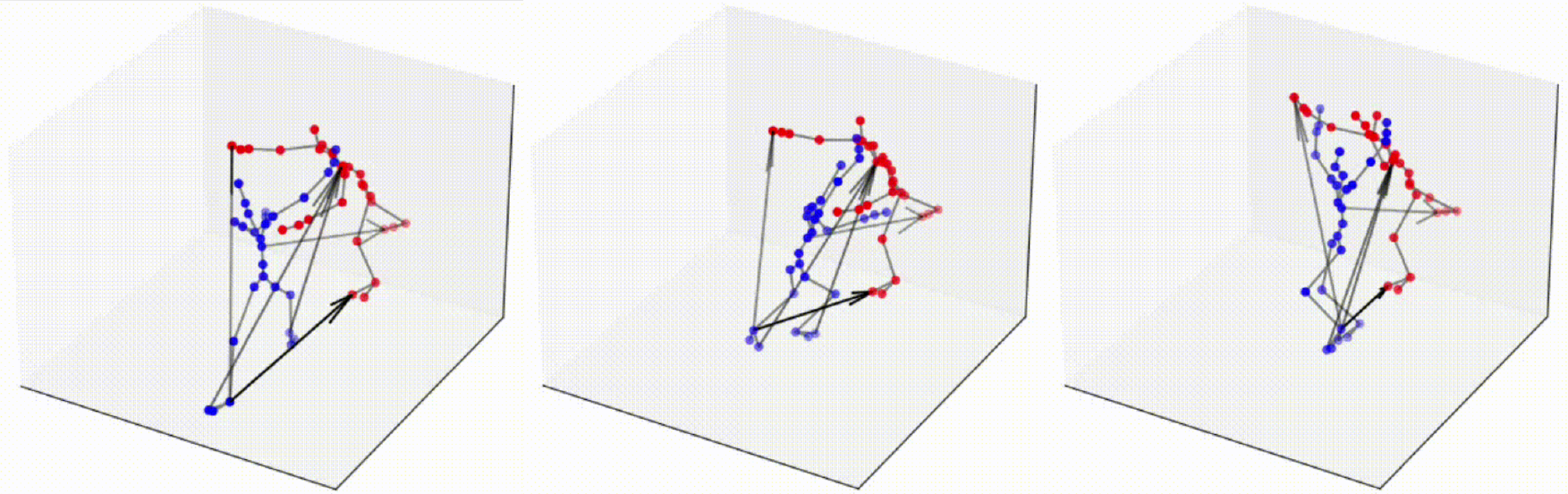}
            \caption{A directed example showcasing more complex movement for 10 sampled joints. Notably the model captures how the foot motion of one dancer influences multiple parts of the other dancer's body. This aligns with the performed movement, where the blue dancer's dynamic spin guides the red dancer's response.}
            \label{multi_edge_2}
        \end{subfigure}
        \caption{Examples of multiple edges connected to the same joint.}
        \label{multi_edges}
    \end{figure}

    \item Another recurring pattern is that \textbf{edges frequently form between joints that are in opposition}, as if connected by an invisible elastic band. These joints appear to be stretching or pulling apart, suggesting that the model is particularly sensitive to moments of tension between dancers (Figure \ref{opposition}). This is especially relevant in the context of duet choreography, where push and pull dynamics are fundamental to many partnering techniques.

    For example, if one dancer extends their arm forward while their partner leans away, the model often identifies a connection between these two points, even if there is no direct contact. This suggests that the model is not only detecting explicit force transfers (such as corresponding body parts in contact) but also implicit movement dependencies, where one dancer’s action influences the other's balance, trajectory, or momentum.

    \begin{figure}[h]
        \centering
        \includegraphics[width=0.45\textwidth, height=2.5cm]{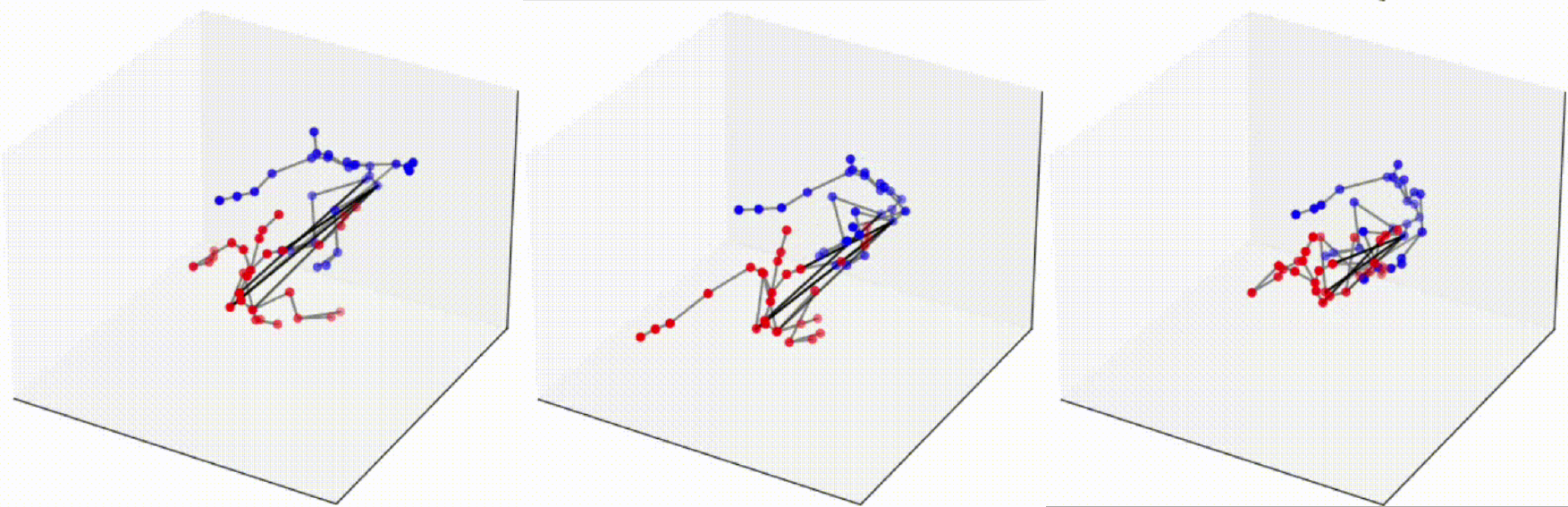}
        \caption{Undirected example of connections within opposition tendencies. It shows multiple connections between the lower torso of both dancers, first leaning in opposite directions and then gravitating toward each other, illustrating the full range of the stretched-string analogy.}
        \label{opposition}
    \end{figure}

    This result aligns closely with our dancers' choreographic intuitions and training, in which movement is often defined by how tension and release are negotiated between partners. It also suggests that the model is learning a representation of movement that extends beyond surface-level trajectory following.

\end{itemize}

\section{Discussion}

\subsection{Creative Implications}

    This work offers a new lens for dancers to analyze their movements and interactions, revealing patterns and relationships they may not consciously recognize. By mapping inter-dancer connections, it provides insights into habitual tendencies, asymmetries, and unconscious influences in movement. For duets, this tool makes visible the often intuitive exchanges between partners — subtle weight shifts, spatial negotiations, and reactive gestures that shape their interaction. Seeing these dynamics mapped out allows dancers to refine their awareness and explore new choreographic possibilities.

    This approach also reimagines dance as a dynamic graph in which dancers are linked by evolving connections. This perspective could inform interactive performances, projecting real-time movement-based visualizations onto a stage. Additionally, it holds potential for dance pedagogy and interdisciplinary research, bridging choreography with biomechanics, for example. By exposing hidden layers of movement, this work expands both artistic and analytical possibilities for dancers and technologists.

\subsection{Future Directions}

    Despite these promising proof-of-concept results, several improvements are needed to enhance model performance and usability. Key areas for future work include:

    \begin{itemize}
        
        \item \textbf{Data expansion and quality:} Regardless of the use of data augmentation through duet rotation, the small dataset size made training difficult. Also the pipeline used to extract 3D poses has room for improvement. Even in original sequences, the dancers’ joints are shaky and often poorly approximated, leading to random (unrealistic) movements. Moreover, normalization of both dancers was removed to preserve relative movement, but it was realized too late that a new layer normalizing their joint movement should have been added - having dancers in different parts of space caused confusion for the models.

        \item \textbf{Architecture exploration:} While many versions of the NRI variant were implemented and tested, the final version is still far from the potential we saw from the strong results achieved by the original version. Furthermore, alternative architectures like transformers \cite{vaswani2023attentionneed} could enhance temporal modeling and better align with modern architectures.

        \item \textbf{Model validation:} Extensive qualitative analysis was conducted through various training sessions, experiments, and even scenario simplifications, but the study did not incorporate a detailed quantitative evaluation beyond learning curve and MSE analysis. Additional metrics and comparative tests with different parameters - such as the number of reconstructed frames - were not much explored and could provide further insights.

        \item \textbf{Processing speed:} Some parts of the final architecture are suboptimal custom implementations. Batches are replaced with sequential operations at several points in the pipeline - node and edge representation conversions, edge sampling for each sequence in a batch, and in the GRNN. As a result, a training cycle with just a few dozen epochs can take an entire day, which is not ideal for scaling.

        \item \textbf{Interaction with dancers:} Since this project sits at the intersection of art and technology, more direct interaction with artists is essential. With a more refined version of the model, it would be ideal to present the tool to the dance community and observe how dancers use the tool in their own partnering studies.
        
    \end{itemize}

\section{Conclusion}

    This work presented an investigation into how AI-driven methods can enhance our understanding of partnered dance by focusing on revealing inter-dancer connections. We paired an open-source 3D pose extraction pipeline with a custom pre-processing stage to capture dancers’ movements and then trained an NRI-based architecture augmented with GCN modules and RNN structure to reveal important edges between the dancers. Though the model only processed a subset of the full-body kinematics of the dancers, it demonstrated a capacity to uncover interesting dance dynamics such as high-confidence inter-joint dependencies that align with choreographic intuition, the presence of key movement hubs that serve as central points of influence within the duet structure and recurring patterns of tension and release.

    While far from comprehensive, our work explores the potential for AI to augment our creative understanding of duet dynamics in choreographic settings. From connections that focus on “push and pull” forces between dancers to more intricate patterns of mutual influence, this project shows the value of graph-based approaches in modeling collaborative creative frameworks.

\section{Author Contributions}

    Authors one and two jointly developed the pose extraction and data processing pipelines. Author one designed and implemented the machine learning methodology and led the paper manuscript writing. Throughout the process, authors three and four provided technical and artistic supervision.

\section{Acknowledgements}
The authors would like to thank the HumanAI program and Google Summer of Code for supporting this work. 

\bibliographystyle{iccc}
\bibliography{iccc}

\end{document}